\documentclass[conference]{IEEEtran}
\IEEEoverridecommandlockouts
\usepackage{cite}
\usepackage{amsmath,amssymb,amsfonts}
\usepackage{algorithmic}
\usepackage{graphicx}
\usepackage{textcomp}
\usepackage{xcolor}

\usepackage[dvipsnames,table]{xcolor}
\usepackage{booktabs,multirow,cases,float}
\usepackage[caption=false,font=footnotesize]{subfig}
\usepackage{url}

\definecolor{red1}{RGB}{220,20,1}
\definecolor{green1}{RGB}{26,153,25}
\definecolor{cvprblue}{rgb}{0.21,0.49,0.74}

\def\BibTeX{{\rm B\kern-.05em{\sc i\kern-.025em b}\kern-.08em
    T\kern-.1667em\lower.7ex\hbox{E}\kern-.125emX}}
\begin{document}

\title{Fractal-Informed Spatial-Frequency Adaptation for Texture Generation in Natural Image Compression}

\author{\IEEEauthorblockN{
Lingyu Zhu\IEEEauthorrefmark{1},
Baoliang Chen\IEEEauthorrefmark{1},
Xiangrui Zeng\IEEEauthorrefmark{1}, 
Chengyan Jiang\IEEEauthorrefmark{1}, 
Liang Xie\IEEEauthorrefmark{2}, and
Shiqi Wang\IEEEauthorrefmark{1}}
\IEEEauthorblockA{\IEEEauthorrefmark{1}City University of Hong Kong, Hong Kong\\
\IEEEauthorrefmark{2}Guangdong University of Technology, Guangdong}}

\maketitle

\begin{abstract}
Generative image compression aims to reconstruct realistic images at low bitrates by balancing the trade-off among rate, distortion, and realism. 
However, recovering plausible textures while preserving reconstruction fidelity remains challenging, particularly under limited bandwidth.
In this paper, we propose a diffusion-informed generative image compression framework that exploits frequency-domain representations and diffusion knowledge to improve texture reconstruction. 
Specifically, we approach the rate-distortion-realism trade-off from three complementary perspectives.
First, motivated by the close relationship between image textures and frequency characteristics, we develop a \textbf{F}ractal \textbf{F}requency-\textbf{A}ware \textbf{B}and \textbf{I}mage \textbf{C}ompression (\textbf{FFAB-IC}) network to capture directional frequency components in natural images. 
By integrating fractal band operations into nonlinear neural transformations, the network preserves essential image information while suppressing redundant details. 
Second, we incorporate diffusion knowledge into the encoder and introduce iterative diffusion refinement into the decoder, enabling the recovery of realistic textures under constrained bitrates. 
Third, we introduce frequency- and content-aware regularization to jointly exploit spatial structures and frequency characteristics during optimization. 
Extensive quantitative and qualitative experiments demonstrate that our method achieves a superior rate-distortion-realism trade-off, delivering higher realism at comparable distortion than existing approaches.
\end{abstract}


\section{Introduction}
\label{sec:intro}
In recent years, learned image compression (LIC) methods have gained considerable attention, showing superior performance compared to traditional codecs~\cite{sullivan2012overview}.
Nonetheless, distortion-focused convolutional neural network (CNN) compression approaches~\cite{minnen2018joint, zhu2024learned} and transformer-based methods~\cite{zhu2022transformer, li2024frequencyaware} solely optimize the rate-distortion trade-off, leading to visually unpleasing results at low bitrates.
These efforts have yielded ``artifact-laden" results, characterized by issues such as blurring and excessive smoothing.
Simply put, these results are ``correct" in the pixel optimization domain, but they fail to align with ``human perception" and ``semantic consistency" due to the absence of perceptual supervision~\cite{johnson2016perceptual} and adversarial supervision~\cite{goodfellow2014generative} within the optimization space.
To prioritize visually pleasing consistency, pioneering researchers explore perceptual-focused learned compression methods~\cite{agustsson2019generative, muckley2023improving, chen2024beyond, chen2025pleno} that leverage generative adversarial networks (GANs) to improve the perceptual quality of output.
The release of diffusion model (DM) techniques~\cite{ho2020denoising, song2020denoising, rombach2022high} has opened up new opportunities.
It has shown remarkable performance in various image tasks, such as image super-resolution \cite{saharia2022image} and image enhancement \cite{lu2024diffusion}, \textit{etc.}
The primary success of these algorithms in low-level tasks can be attributed to the utilization of pre-trained diffusion models

\begin{figure*}[h]
\hsize=\textwidth 
\centering
\includegraphics[width=1.0\textwidth]{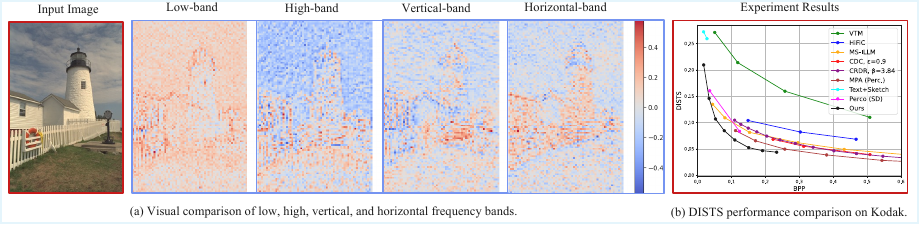}
\caption{\textbf{Motivation and superiority.} 
The visual results show the directional frequency bands captured through the designed FFAB block.
Each frequency band demonstrates distinct characteristics, allowing for an intuitive understanding of the inherent fractal patterns within natural images.
(b) Our FFAB-IC network demonstrates a significant performance improvement on the Kodak dataset, as evidenced by the DISTS value.}
\label{fig:motivation}
\end{figure*}

The generative image compression technique fundamentally relies on signal synthesis, emphasizing the importance of explicitly capturing the frequency band dependencies of images to recover texture details.
Although existing learned generative image compression methods have shown good performance~\cite{li2024towards, mentzer2020high, yang2024lossy}, they often fail to interpret the distinct characteristics of the frequency, providing an intuitive understanding of the inherent fractal patterns in wavelets~\cite{edwards1991discrete}, steerable filters~\cite{freeman1991design}, wiener filters~\cite{paul2024f2former}, and ringlets~\cite{do2005contourlet}, together with related processing methods~\cite{pan2022fast, li2024frequencyaware, zhu2022enlightening}.
This exploration inspires us to leverage frequency techniques and their diverse applications, paving the way for a new paradigm in the rate-distortion-realism tradeoff. 
Our aim is to push the boundaries of achievable distortion and realism in generative image compression.
In the realm of image compression, the diffusion model governs the generation process by conditioning constrained latent variables, enabling a more nuanced approach to recovering lost details.

However, directly applying pre-trained diffusion for learned generative image compression is not practical, and current generative image compression methods still face several challenges, which we summarize as follows,
\begin{itemize}
    \item Despite the success of pre-trained stable diffusion, the inability to directly utilize pre-trained stable diffusion models in codec design stems from their fundamental architectural differences and operational constraints. 
    As such, bridging the gap between these two domains requires innovative strategies that can leverage the strengths of stable diffusion while addressing the limitations imposed by codec functionalities. 

    \item Generative models prioritize producing high-quality content without bitrate supervision, while compression techniques aim to efficiently represent and reconstruct images within limited bitrates following Shannon’s information theory~\cite{shannon1959coding}. 
    The inherent conflict between the stochastic nature of the diffusion process and the constraints imposed by bitrate requirements presents a significant issue.

\end{itemize}
To solve the above-mentioned challenges, we aim to develop the generative compression architecture that bridges the two regimes, effectively harmonizing the rate-distortion-realism trade-off between diffusion knowledge and compression techniques.
The proposed FFAB-IC network integrates generative prior with compression techniques by designing the Fractal Frequency-Aware Band (FFAB) block, which has two notable characteristics.
\textbf{Firstly}, it connects frequency band recovery theory under a rate constraint, allowing it to efficiently capture the directional frequency components within the natural image through isotropic and anisotropic window attention.
That is, the majority of image information is contained in the low-frequency component, which serves as the basic element, while a smaller portion of texture details is found in the high-frequency components, which helps capture potential hierarchical dependencies among frequency band features.
\textbf{Secondly}, the designed FFAB block integrates a generative prior to the encoding and decoding processes, thus enhancing the performance of image realism.
Connecting it within a unified block enhances frequency band learning under rate constraints, resulting in superior texture modeling capacity. 
Additionally, frequency-aware and content-aware supervision is employed to regularize network training to enhance the modeling capability.
This approach ensures consistency in the extracted features, promoting a more robust and effective representation during the compression process.
The main contributions of our work are as follows,
\begin{itemize}
\item We propose an innovative pipeline that integrates generative perceptual knowledge for image compression, tackling the \textbf{critical challenge} of balancing rate, distortion, and realism.
This method addresses the inherent conflict between the stochastic nature of the diffusion process and the deterministic requirements of image compression.

\item We design a \textbf{FFAB-IC} network, which integrates and leverages the designed FFAB block to capture directional frequency components of natural images, enabling adaptive latent representation in an end-to-end manner.

\item We employ the frequency-aware and content-aware \textbf{supervision} constraints to regularize the training of the FFAB-IC model. 
This constraint modulates the 
compression network to align with the diffusion perception space.

\item We conduct \textbf{extensive experiments}  to validate the proposed FAFB-IC model. 
The results demonstrate that it significantly outperforms various baseline models across multiple public benchmarks. 
Furthermore, it excels compared to competitors in terms of visual perception quality in real-world scenarios.
\end{itemize}

\begin{figure*}[t]
\centering
\includegraphics[width=0.85\linewidth]{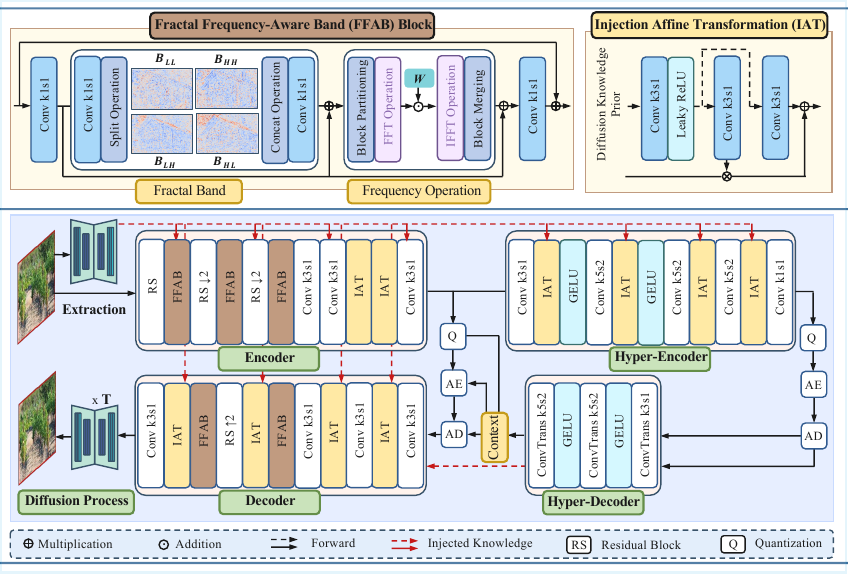}
\caption{Overview of the proposed Fractal Frequency-Aware Band Image Compression (FFAB-IC) Framework.
Our FFAB-IC framework leverages a pre-trained Stable Diffusion, integrating generative prior knowledge with Fractal Frequency-Aware Band information to enhance contextual understanding in two key aspects:
\textbf{(a)} Feature-Level Operation: the framework enables window attention interactions across low, high, vertical, and horizontal bands. 
This allows for effective integration between generative features and image features in the representation space.
\textbf{(b)} Optimization-Level Guidance: we introduce frequency domain constraints into the computation of content representation.}
\label{fig: framework}
\end{figure*}

\section{Methodology} 
\noindent \textbf{Motivation.} 
The strong generative capabilities of stable diffusion inspire us to investigate a distortion-realism trade-off approach for image compression. 
Two drawbacks hinder existing generative compression methods in real-world scenarios.
Firstly, retrieving high-frequency details becomes challenging when inputs experience significant degradation at low bitrates through discriminative learning~\cite{wu2024latent}.
Secondly, generative compression methods often neglect the significance of frequency band information, resulting in missing texture details. 
As illustrated in Fig.~\ref{fig: framework}, the architecture of the proposed FFAB-IC network is presented to address this challenging issue.
The core design is built upon the FFAB block, enabling non-linear latent representation mapping for better compression. 
The FFAB block introduces frequency decomposition window attention in a fractal structure, band split, band refinement, and band fusion to capture low-frequency $B_{LL}$, high-frequency $B_{HH}$, vertical $B_{HL}$, and horizontal components $B_{LH}$, along with a frequency modulation technique for the adaptive adjustment of these components.

\subsection{Preliminaries}
\noindent \textbf{Preliminary on Compression Model.} 
A neural image compression method comprises three fundamental components~\cite{mentzer2020high}: an encoder \( E \), a decoder \( G \), and an entropy model \( P \).
Specifically, the encoder \( E \) transforms an input image \( \boldsymbol{x} \) into a quantized latent representation 
\( \boldsymbol{y} = E(\boldsymbol{x})\).
The decoder \( G \) takes the latent representation \( \boldsymbol{y} \), generating a reconstruction of the original image
\( 
\boldsymbol{x}^{\prime} = G(\boldsymbol{y}) \).
The total objective is to minimize the rate-realism trade-off~\cite{cover1999elements} as follows,
\begin{equation}
L_{\text{total}}=\mathbb{E}_{\boldsymbol{x} \sim p_{\boldsymbol{X}} }
\left[\lambda R(\boldsymbol{y}) + D \left(\boldsymbol{x}, \boldsymbol{x}^{\prime}\right)\right]. 
\end{equation}
The rate is estimated through the cross-entropy measure, defined as \( R(\boldsymbol{y}) = -\log P(\boldsymbol{y}) \). 
Furthermore, \( D(\boldsymbol{x}, \boldsymbol{x}^{\prime}) \) represents the metric that is used to evaluate the realism of the decoded image.

\noindent \textbf{Preliminary on Diffusion Model.} 
Diffusion model~\cite{ho2020denoising} is the generative framework that systematically adds noise to data and subsequently learns to reverse this process for sample generation. 
This pipeline begins with a real data distribution $\boldsymbol{x} \sim p_{\boldsymbol{X}}$, where an initial data point \(\boldsymbol{x}_0\) is progressively transformed into noise across multiple time steps \(t\).
In each step, the data transitions toward randomness, with the noise level determined by a time-varying parameter \( \beta_t \), which increases over time.
Eventually, after sufficient iterations, the data distribution approaches a standard normal distribution, represented as \( \boldsymbol{x}_t \sim \mathcal{N}(0, \mathbf{I}) \).
To generate new samples, the model reverses this process by reconstructing the original data \(\boldsymbol{x}_0 \) from the noisy input \( \boldsymbol{x}_t \). 
This is accomplished by learning the denoising function with a neural network, parameterized by \( \theta \), which outputs the clean data mean \( \mu_\theta(\boldsymbol{x}_t, t) \) and the associated noise variance \( \sigma^2(t)\).

\begin{figure}[h]
\centering
\includegraphics[width=0.8\linewidth]{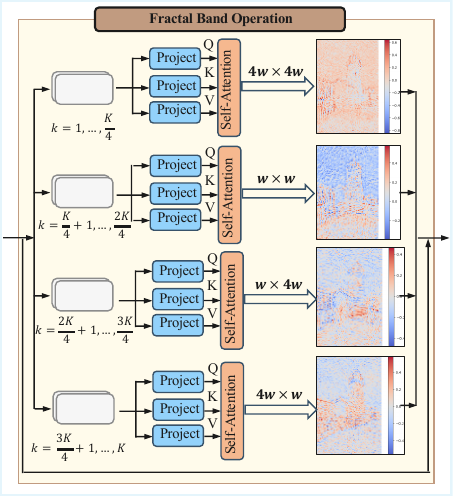}
\caption{We present the Fractal Frequency-Aware Band (FFAB) Block.
This block utilizes the Fast Fourier Transform (FFT) to convert content variables from the spatial domain to the frequency domain.}
\label{fig: fractal band operation}
\end{figure}

\begin{figure*}[t]
\centering    
\subfloat[Bpp $|$ FID $|$ DISTS] 
{\includegraphics[width=0.245\linewidth]{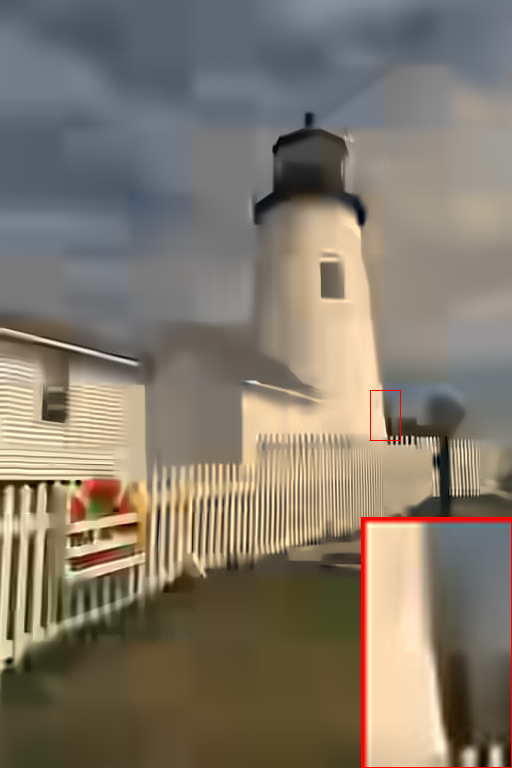}} \hskip0.2em
\subfloat[0.1322 $|$ 53.73 $\downarrow$ $|$ 0.0843 $\downarrow$] 
{\includegraphics[width=0.245\linewidth]{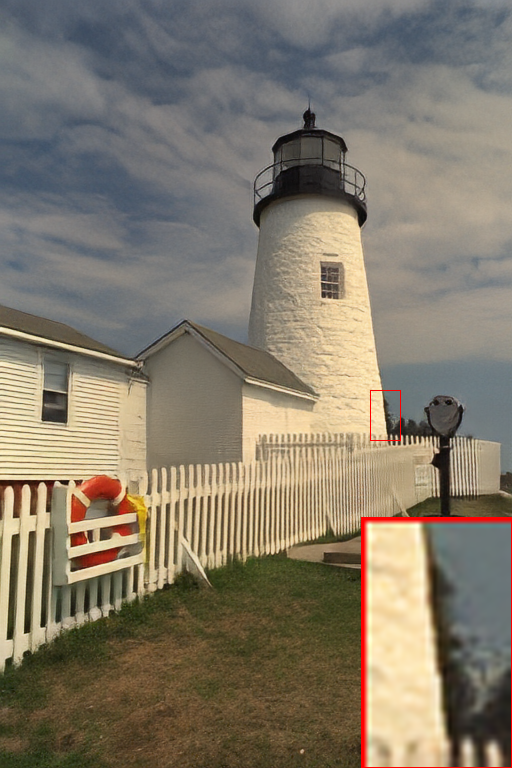}} \hskip0.2em  
\subfloat[0.1354 $|$ 39.36 $\downarrow$ $|$ 0.0676 ] 
{\includegraphics[width=0.245\linewidth]{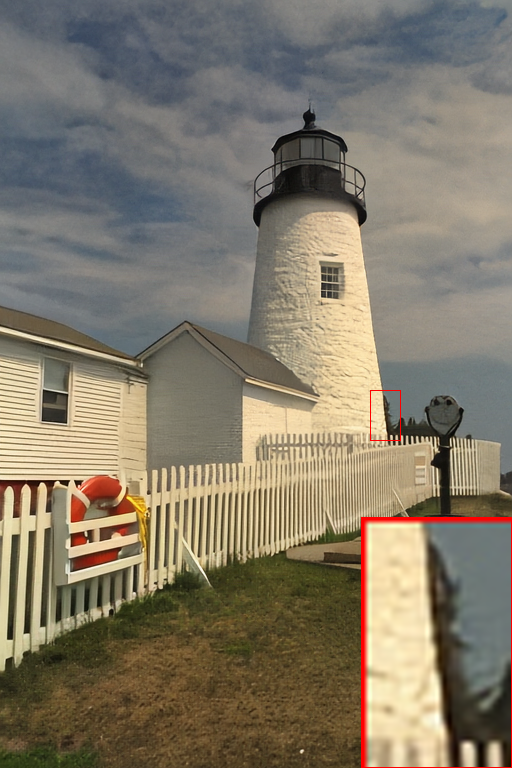}} \hskip0.2em
\subfloat[0.1465 $|$ 22.48 $\downarrow$ $|$ 0.0454 $\downarrow$] 
{\includegraphics[width=0.245\linewidth]{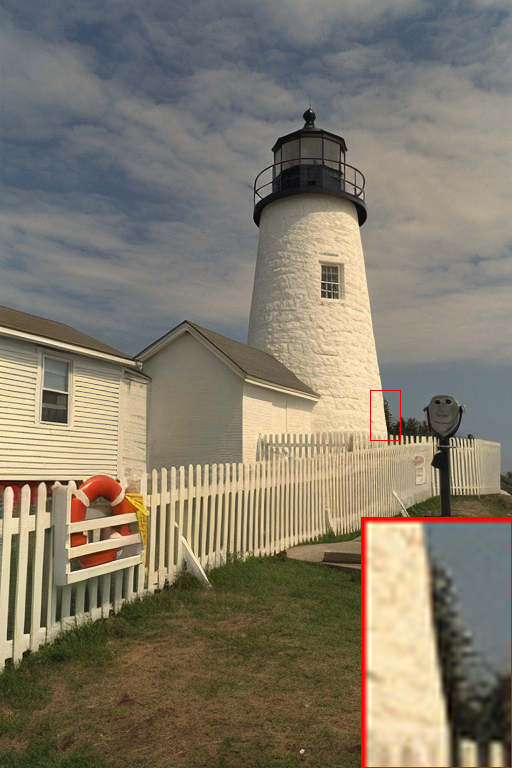}} 
\setlength{\abovecaptionskip}{1mm}
\setlength{\belowcaptionskip}{-3mm}
\caption{
Visual comparison of compression methods (a) VTM~\cite{bross2021overview}, (b) HiFiC~\cite{mentzer2020high}, (c) CRDR~\cite{iwai2024controlling}, and (d) Ours.
Our method could effectively maintains fine details and improve overall image quality, as evidenced by the clarity in zoomed-in areas.
}
\label{fig: visual comparison}
\end{figure*}

\noindent \textbf{Fractal Frequency-Aware Band Learning.}
Fractal Band Learning represents a promising paradigm for analyzing signals across different sub-bands, enabling the capture of intricate details~\cite{singh2014sub}.
Observations from \cite{li2024frequencyaware} inspire us that typical self-attention acts as a low-pass filter. 
Utilizing the isotropic and anisotropic windows could capture the directional frequency components within natural images, which is seldom investigated in fractal band learning.
As such, we focus on analyzing specific aspects of band recovery characteristics writing into the compact bitstream. 
The signal is reconstructed using a two-step process.
\textbf{1)}. The residual block \( F_{\mathrm{RB}} \) generates a new band signal \( f_k \) based on the previous band estimator \( f_{k-1} \).
\textbf{2)}. A summation combines the new band signal (residue) \( F_{\mathrm{RB}}(f_{k-1}) \) with the previous band estimator \( f_{k-1} \) as follows,
\begin{equation}
f_k=F_{\mathrm{RB}}\left(f_{k-1}\right)+f_{k-1},
\end{equation}
wherein \( F_{\mathrm{RB}}(\cdot) \) denotes the operation of the residual block and $f_k$  denotes the four band features in the frequency domain (\textit{i.e., isotropic and anisotropic features}).
Compared to existing methods in \cite{singh2014sub, yang2017deep}, our uniquely designed block effectively separates the feature signal into multiple components with distinct intrinsic frequencies, namely low-frequency, high-frequency, vertical, and horizontal bands. 
These components, referred to as $\left\{B_{LL},  B_{HH}, B_{HL}, B_{LH}\right\}$, are modeled individually.
That is, the majority of image information is contained in the low-frequency component, which serves as the basic element, while a smaller portion of texture details is found in the high-frequency components, which helps capture potential hierarchical dependencies among frequency band features
We visualize the corresponding bands for intuitive understanding, as shown in Fig.~\ref{fig:motivation} (a).

\subsection{Generative Compression Framework}
\noindent \textbf{Overall.}
The generative feature derived from the pre-trained stable diffusion $\boldsymbol{y}_d = F_d(\boldsymbol{x})$ is extracted from raw image $\boldsymbol{x}$ for utilization.
Similarly, the analysis transforms $F_a(\cdot)$ maps $\boldsymbol{x}$ to a latent representation $\boldsymbol{y}$.
The hyper-information is derived using a pair of hyper encoder $H_a(\cdot)$ and hyper decoder $H_d(\cdot)$.
$Q(\cdot)$ denote the quantization operator.
The encoding (Enc.) process can be summarized as follows,
\begin{numcases}{\text{Enc.}} 
\boldsymbol{y}=F_a(\boldsymbol{x}, \boldsymbol{y}_d), \quad \hat{\boldsymbol{y}}=Q(\boldsymbol{y}), \label{equ:y} \\
\boldsymbol{z}=H_a(\boldsymbol{y}, \boldsymbol{y}_d), \quad \hat{\boldsymbol{z}}=Q(\boldsymbol{z}).
\label{equ:z}
\end{numcases}

\begin{figure*}[t]
\centering    
\subfloat[Bpp $|$ FID $|$ DISTS] 
{\includegraphics[width=0.45\linewidth]{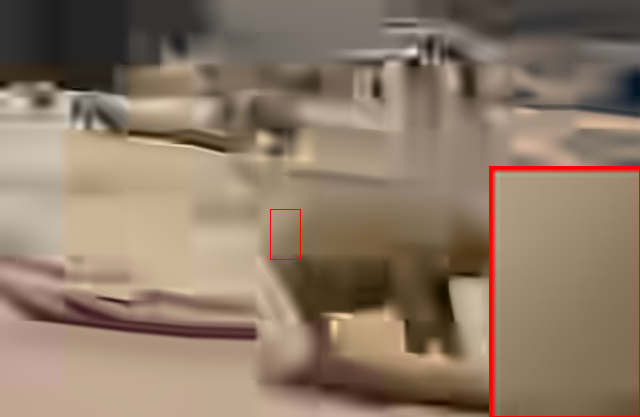}} \hskip0.2em
\subfloat[0.1805 $|$ 31.50 $\downarrow$ $|$ 0.1139 $\downarrow$] 
{\includegraphics[width=0.45\linewidth]{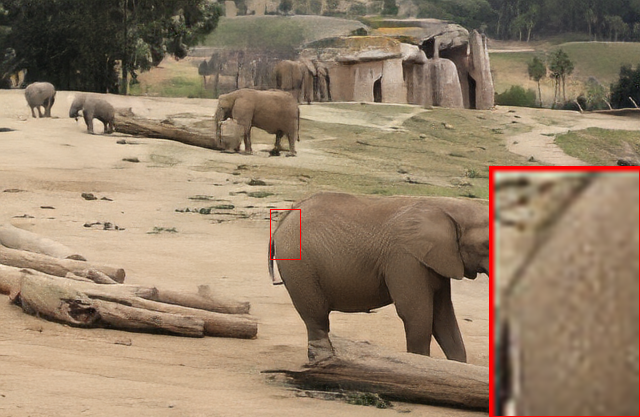}} \hskip0.2em


\subfloat[0.1129 $|$ 36.57 $\downarrow$ $|$ 0.1161 $\downarrow$] 
{\includegraphics[width=0.45\linewidth]{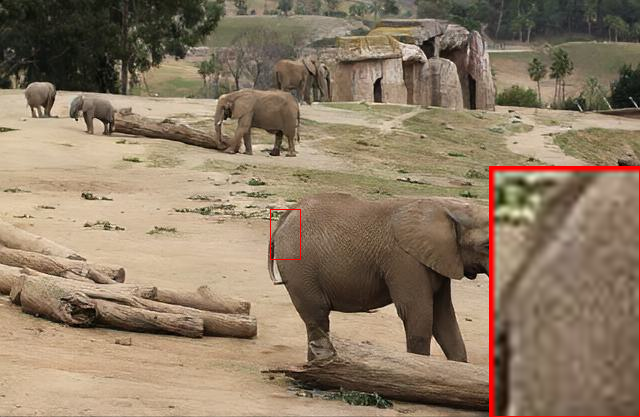}} \hskip0.2em
\subfloat[0.1411 $|$ 19.89$\downarrow$ $|$ 0.0564 $\downarrow$] 
{\includegraphics[width=0.45\linewidth]{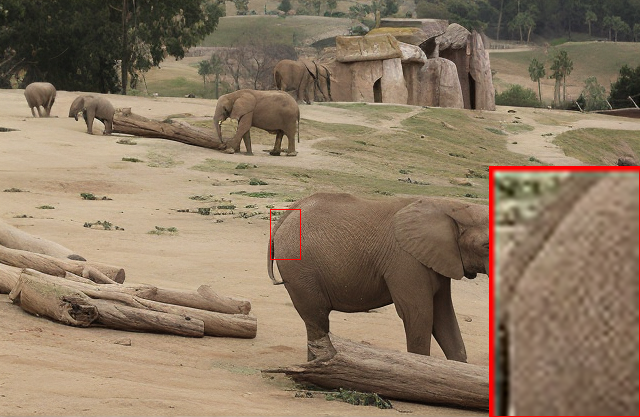}} 
\caption{
Visual comparison of compression methods (a) VTM~\cite{bross2021overview}, (b) HiFiC~\cite{mentzer2020high}, (c) MS-ILLM \cite{muckley2023improving}, (d) CRDR~\cite{iwai2024controlling}, (e) CDC~\cite{yang2024lossy}, (f) Ours.
Our method could effectively preserve fine details and improve overall image quality, as evidenced by the clarity in zoomed-in areas.
}
\label{fig: visual comparison 5}
\end{figure*}

Given the transmitted $\hat{\boldsymbol{y}}$ and $\hat{\boldsymbol{z}}$, we further use the information extraction function $H_d^{w}(\hat{\boldsymbol{z}})$, which has the same architecture as the hyper-decoder. 
Following the framework of the previous channel-wise autoregressive entropy model \cite{he2022elic}, we divide $\boldsymbol{y}$ into 10 uneven slices to enhance the encoding of subsequent slices. 
We  decode $\hat{\boldsymbol{y}}$ into a content representation $\boldsymbol{z}_c$.
Then the content representation $\boldsymbol{z}_c$ is further decoded in the subsequent image reconstruction stage using stable diffusion $G^{t}$.
This decoding (Dec.) process can be expressed as follows,
\begin{numcases}{\text{Dec.}} 
\boldsymbol{z}_c=F_d(\hat{\boldsymbol{y}}, H_d^{w}(\hat{\boldsymbol{z}})), \\
\hat{\boldsymbol{x}}=G^{t}(\boldsymbol{z}_c),  t = 1, 2, \ldots, T.
\end{numcases}

\begin{table}[t]
\centering
\caption{
BD-rate (\%) comparison is calculated on the Kodak~\cite{kodak2013}, DIV2k~\cite{agustsson2017ntire} and MS-COCO 30K~\cite{lin2014microsoft} datasets, wherein the HiFiC is the anchor. 
[Key: \textbf{\textcolor{red1}{Best}}, \textcolor{green1}{\underline{Second Best}}, $\downarrow$ ($\uparrow$): Smaller (larger) values denote better performance].
``-" indicates that the range is outside the scope of calculating the BD-rate.
}
\scalebox{0.8}{
\begin{tabular}{l|cc|cc|cc} 
\hline \hline
\multirow{2}{*}{\textbf{Methods}}   & \multicolumn{2}{c|}{\textbf{Kodak}}  & \multicolumn{2}{c|}{\textbf{DIV2K}}  & \multicolumn{2}{c}{\textbf{MS-COCO 30K}}  \\
& \textbf{DISTS} $\downarrow$  & \textbf{FID} $\downarrow$ & \textbf{DISTS} $\downarrow$    & \textbf{FID} $\downarrow$  & \textbf{DISTS} $\downarrow$    & \textbf{FID} $\downarrow$ \\\hline
\rowcolor{cvprblue!10}
VVC~\cite{bross2021overview}                      & -        & -       & - & -        & -       & - \\\hline
HiFiC \cite{mentzer2020high}       & 0.00\%          & 0.00\%       & 0.00\% & 0.00\%          & 0.00\%       & 0.00\%\\
\rowcolor{cvprblue!10}
MS-ILLM \cite{muckley2023improving}         & -41.52\%           &   -41.43\%     & -41.91\%                      & -38.27\% & +49.26\%       & - \\
CDC \cite{yang2024lossy}                                           & +67.97\%          & -46.39\%           & \textcolor{green1}{\underline{-46.39\%}}                  & -5.34\% & +36.45\%                & -\\
\rowcolor{cvprblue!10}
CRDR  \cite{iwai2024controlling}                               &  -31.88\%             & -13.19\%      & -30.16\%                            & - & -25.88\%                            & -\\ 
MPA \cite{zhang2024allinone}  & \textcolor{green1}{\underline{-64.25\%}}           & \textcolor{green1}{\underline{-51.30\%}}       & -                           & \textbf{\textcolor{red1}{-46.45\%}} & \textcolor{green1}{\underline{-64.13\%}}                          & \textcolor{green1}{\underline{-48.67\%}}\\\hline
\rowcolor{cvprblue!10}
Ours             & \textbf{\textcolor{red1}{-75.50\%}}             & \textbf{\textcolor{red1}{-73.21\%}}         & \textbf{\textcolor{red1}{-57.45\%}}     & \underline{\textcolor{green1}{-40.86\%}}  & \textbf{\textcolor{red1}{-83.09\%}}       & \textbf{\textcolor{red1}{-84.37\%}} \\\hline\hline
\end{tabular}}
\label{tab: BD-Rate on four dataset}
\end{table}

\noindent \textbf{Fractal Frequency-Aware Band (FFAB) Block.}
As shown in Fig.~\ref{fig: fractal band operation}, the designed FFAB block integrates both isotropic and anisotropic window attention mechanisms in the frequency domain. 
The core function lies in the fact that isotropic attention treats all directions equally and is effective in capturing regular patterns, while anisotropic attention enables varying attention patterns based on direction.
Then, the adaptive attention mechanism is designed to encode various patterns into a compact bitstream, facilitating subsequent texture recovery under the rate constraints.
In our experiments, the various window sizes allow the designed FFAB to effectively capture low-frequency, high-frequency, vertical, and horizontal components. 
These window configurations correspond to the respective frequency and directional characteristics.

Specifically, we begin by linearly projecting the latent features into \( K \) heads.
These heads are then evenly divided into four parallel groups, each containing \( \frac{K}{4} \) heads. 
Each group employs a distinct self-attention mechanism, and the outputs from these four groups are concatenated to form the overall output.
Recognizing that the contributions of these frequency components to compression are not equal, we utilize the Fast Fourier Transform (FFT) to convert the content variables from the spatial domain to the frequency domain. 
Specifically, we apply a block-based FFT to the features obtained from a standard feed-forward network (FFN).
Next, we introduce a learnable filter matrix \( \boldsymbol{W} \) to suppress or amplify the various frequency components selectively.
A frequency-modulation feed-forward network then modulates the decomposed components to reduce potential redundancy across different frequency components. 
This process can be formally expressed as follows,
\begin{equation}
\operatorname{FFAB}(\boldsymbol{X}) = \text{Concat}\left[\text{head}_1, \ldots, \text{head}_K\right] \boldsymbol{W},
\end{equation}
wherein 
\begin{equation}
\text{head}_k = 
\begin{cases}
\mathrm{B}^{k}_{LL} & \text{for } k = 1, \ldots, \frac{K}{4} \\
\mathrm{B}^{k}_{HH} & \text{for } k = \frac{K}{4} + 1, \ldots, \frac{2K}{4} \\
\operatorname{B}^{k}_{HL} & \text{for } k = \frac{2K}{4} + 1, \ldots, \frac{3K}{4}\\
\operatorname{B}^{k}_{LH} & \text{for } k = \frac{3K}{4} + 1, \ldots, K
\end{cases}
\end{equation}
and $\boldsymbol{W}$ is the projection matrix that implements interaction between different frequency components.

\noindent \textbf{Injection Affine Transformation (IAF).}
As shown in Fig.~\ref{fig: framework}, we introduce an injection affine transformation to incorporate generative knowledge $\boldsymbol{y}_{d}^{i}$ into the encoder and decoder of our compression pipeline. 
This mechanism facilitates the integration of stable diffusion knowledge prior to adaptively modifying the compressed features, effectively merging spatial and frequency information. 
Herein, the final feature leverages trainable scale (expressed as Eq.~\eqref{equ:gamma}) and shift (expressed as Eq.~\eqref{equ:beta}) factors to refine the learnable distribution.
\begin{numcases}{\text{}} 
\gamma = W_{\text{gamma}}\ast \boldsymbol{y}_{d} + b_{\text{gamma}}
\label{equ:gamma},\\
\beta = W_{\text{beta}} \ast \boldsymbol{y}_{d} + b_{\text{beta}},
\label{equ:beta} 
\end{numcases}
As a result, we apply equivalent transformations to these features, balancing the distribution of activations and weights and making the model more suitable for neural feature compression. 
This simple process is expressed as follows,
\begin{equation}
\boldsymbol{y} = \boldsymbol{y} \cdot (1 + \gamma) + \beta.
\end{equation}

\subsection{Optimization}
\noindent The total loss for the proposed generative image compression method is defined as,
\begin{equation}\label{total loss}
L_{\text {Total }}=\underbrace{\lambda_1 L_{\text {rate }}}_{\text {rate term}} + \underbrace{\lambda_2 L_{\text {spatial}}}_{\text {spatial term}} + \underbrace{ \lambda_3 L_{\text {frequency}}}_{\text {frequency term}} + \underbrace{ \lambda_4 L_{\text {noise}}}_{\text {diffusion term}}
\end{equation}
wherein $\lambda_{1}$, $\lambda_{2}$, $\lambda_{3}$ and $\lambda_{4}$ represent the weights assigned to the rate constraint, spatial content, frequency content, and diffusion loss, respectively.

\noindent \textbf{Rate Constraint.}
We utilize the rate loss $L_{\text {rate}}$ to estimate the rate performance, defined as follows:
\begin{equation}
\begin{aligned}
L_{\text{rate}}= & R(\hat{\boldsymbol{y}}) + R(\hat{\boldsymbol{z}}), \\
\end{aligned} 
\end{equation}
where $R(\cdot)$ denotes the bit rates of the latent representation.

\noindent \textbf{Spatial Domain Constraint.} 
In the proposed FFAB-IC network, we employ the spatial domain constraint to ensure that the content variables align with the diffusion content space, thereby providing essential constraints for optimization.
This process could be expressed as follows,
\begin{equation}
L_{\text{spatial}}=\left\|\boldsymbol{z}_c-F_d(\boldsymbol{x})\right\|^2
\end{equation}

\noindent \textbf{Frequency Domain Constraint.}
Motivated by distance measurement in the frequency domain \cite{li2024osmamba}, we further employ the FFT operation to convert the content variables from the spatial domain to the frequency domain. 
We then compare the differences in the amplitude component \( A \) and the phase component \( P \) as follows,
\begin{equation}
L_{\text{Frequency}} = \left\| A(\boldsymbol{z}_c) - A(F_d(\boldsymbol{x})) \right\|^2 + \left\| P(\boldsymbol{z}_c) - P(F_d(\boldsymbol{x})) \right\|^2
\label{frequency_equ}
\end{equation}

\noindent \textbf{Diffusion Term.} 
To estimate the noise, the introduction of the external content condition 
$z_c$.
\begin{equation}
L_{\text{noise}}=\mathbb{E}_{\boldsymbol{z}_0, t, \epsilon, \boldsymbol{z}_c}\left\|\epsilon-\epsilon_\theta\left(\boldsymbol{z}_t, t, c, \boldsymbol{z}_c\right)\right\|^2    
\end{equation}
wherein $c$ is the empty text condition.

\section{Experiments}
\subsection{Experiment Settings}
\label{Experiment Settings}

\noindent \textbf{Training Settings}.
All experiments are conducted on a single NVIDIA 3090 GPU. 
The comparison method uses the released code while ensuring the experimental settings are consistent with ours.
We employ Stable Diffusion 2.1-base as the specific implementation and utilize it as the diffusion prior.
Throughout all our experiments, the weights of stable diffusion remain frozen. 
For optimization, we utilize Adam~\cite{kingma2014adam} optimizer with $\beta_1 = 0.9$ and $\beta_2 = 0.999$ and set the learning rate to $10^{-4}$. 
We train the proposed compression method, and the corresponding $\lambda_2$, $\lambda_3$, and $\lambda_4$ are set to 2, 0.0001, and 1, respectively. 
We train six models with $\lambda_1$ selected from \{0.5, 1, 2, 4, 8, 16\} to achieve different bit rates.
We train the base model $\lambda_1=1$ for 300K iterations and fine-tune the other rates for about 50K iterations.
During training, the images are randomly cropped to a size of \( 256 \times 256 \times 3 \).
The training batch size is set to 4.

\noindent \textbf{Dataset.}
The proposed method is trained on the LSDIR~\cite{li2023lsdir} dataset, which contains 84,911 high-quality images.
We adopt three testing datasets to evaluate the effectiveness of the proposed compression method.
\textbf{a) Kodak~\cite{kodak2013}:} This dataset includes 24 images, each sized either 512 x 768 or 768 x 512 pixels.
\textbf{b) DIV2K~\cite{agustsson2017ntire}:} The validation set has 100 high-quality images, each with a resolution of 2000 pixels. 
\textbf{c) MS-COCO 30K~\cite{lin2014microsoft}:} The data set comprises 30,000 images sourced from the MS-COCO 2017 training set, which is used to assess the realism of compression methods, following previous research~\cite{careil2024towards}.

\noindent \textbf{Evaluation Metrics.}
We adopt the Deep Image Structure and Texture Similarity
(\textbf{DISTS})~\cite{ding2020image}, Fréchet Inception Distance (\textbf{FID})~\cite{heusel2017gans} to evaluate the quality of the decompressed images.
For calculating FID, we follow the protocol established in previous works~\cite{mentzer2020high, agustsson2023multi} to ensure a robust quality evaluation. 

\noindent \textbf{Baseline Methods.} 
We compare the proposed method against six representative image compression methods, which include one traditional codec, two GAN-based codecs, one diffusion-based method, and two unified codecs. 
In particular, these include the traditional compression standard VVC Intra (VTM-12.0)~\cite{bross2021overview},
GAN-based compression methods such as HiFiC \cite{mentzer2020high} and MS-ILLM \cite{muckley2023improving},
diffusion-based method include Conditional Diffusion Compression (CDC, $\rho$ = 0.9) \cite{yang2024lossy}.
Additionally, we also evaluate unified models, including MPA (PERC.)~\cite{zhang2024allinone} and CRDR ($\beta$ = 3.84) \cite{iwai2024controlling} that focus on rate-perception performance.

\begin{figure}[t]
\centering    
{\includegraphics[width=0.9\linewidth]{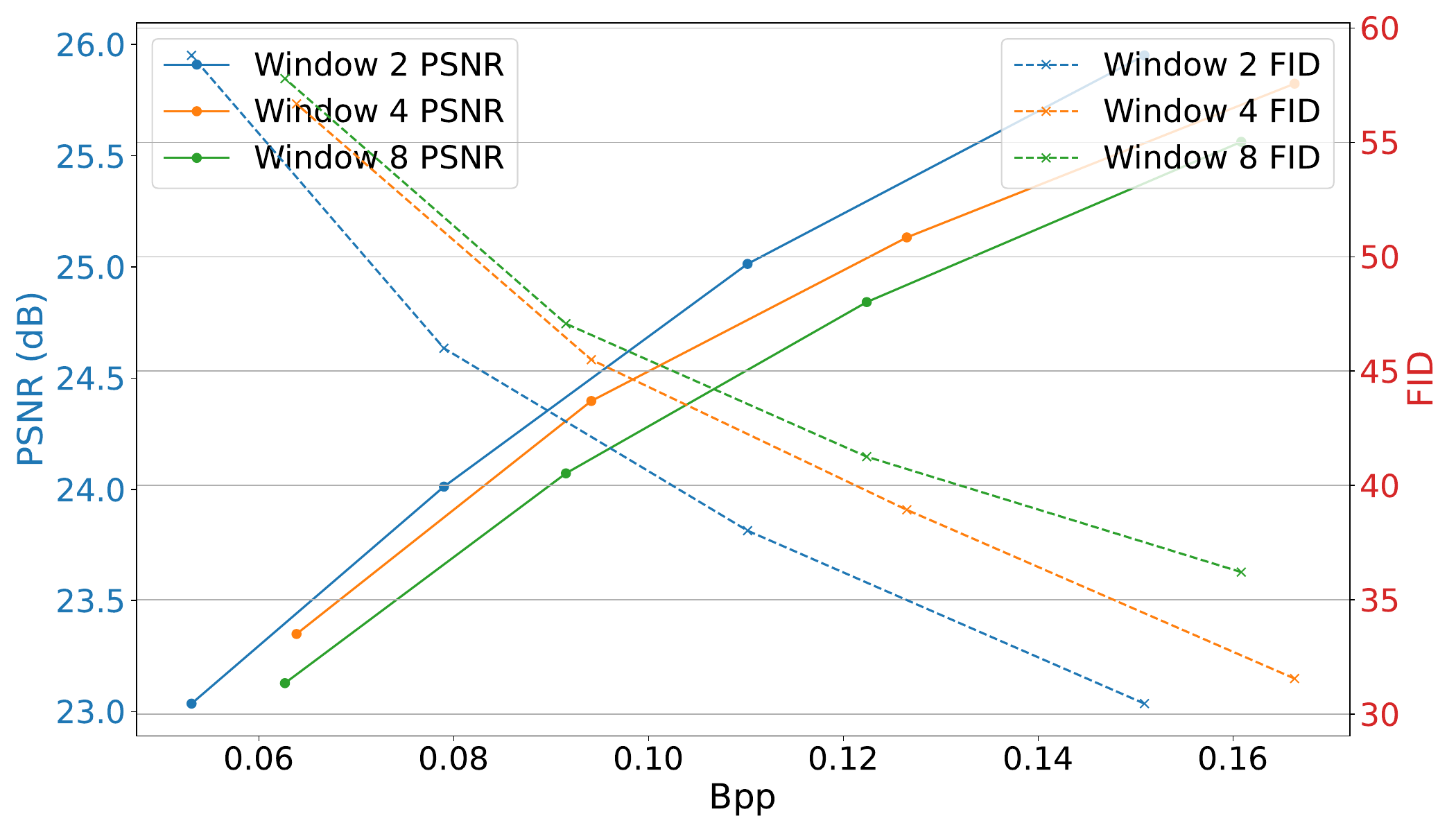}} 
\setlength{\abovecaptionskip}{0mm}
\setlength{\belowcaptionskip}{-4mm}
\caption{
Illustration of the relationship between bits per pixel (Bpp) and the corresponding PSNR and FID values for window 2, 4, and 8 settings.}
\label{fig: FFAB window size}
\end{figure}

\subsection{Experimental Results}
\noindent \textbf{Quantitative Comparisons.}
Tab.~\ref{tab: BD-Rate on four dataset} provide a quantitative comparison between our proposed compression method and other leading methods. 
We utilize HiFiC as the anchor for calculating the BD-rate, allowing us to assess performance improvements across three testing datasets: Kodak, DIV2K, and MS-COCO 30K.
\textbf{Observation:} 
We achieve the BD-rate improvements on the Kodak dataset, with a remarkable performance of -75.50\% in DISTS and -73.21\% in FID, underscoring the ability to preserve image quality while compressing data effectively. 
On the DIV2K dataset, our compression method could obtain DISTS and FID gain by -57.45\% and -40.86\%, respectively.
For the MS-COCO 30K dataset, our method exhibits even more remarkable improvements, achieving -83.09\% in DISTS and -84.37\% in FID, indicating that our model outperforms other methods in complex image scenarios.
\textbf{Analysis:} 
The results demonstrate that our proposed method is a competitively effective learned generative image compression technique. It achieved first place across all three datasets in terms of DISTS and FID scores, highlighting the effectiveness of our generative image compression.

\begin{table}[t]
\centering
\small
\setlength{\abovecaptionskip}{1mm}
\caption{Encoding (Enc.) and decoding (Dec.) time (seconds), as well as BD-rate (FID) (\%) computed on the Kodak dataset \cite{kodak2013} are summarized using an Nvidia GPU 3090, with HiFiC serving as the anchor.
}
\scalebox{0.95}{
\begin{tabular}{l|ccc} \hline\hline
\textbf{Methods} & \textbf{Enc. Time} & \textbf{Dec. Time} & \textbf{BD-rate}   \\\hline
\rowcolor{cvprblue!10}
HiFiC \cite{mentzer2020high}   & 0.598  & 1.374  & 0.00\%    \\
MS-ILLM \cite{muckley2023improving}             &  0.434  &  0.101 &  -41.43\% \\
\rowcolor{cvprblue!10}
CDC ($\rho$ = 0.9) \cite{yang2024lossy}      & 1.087   & 0.008 & -40.15\%  \\ 
CRDR ($\beta$ = 3.84) \cite{iwai2024controlling}                  & 3.269  &0.896 & -13.19\%    \\ 
\rowcolor{cvprblue!10}
MPA (PERC.)~\cite{zhang2024allinone}            &  0.386   &  0.330   &  -51.30\%  \\\hline
Ours      & 0.507    & 0.221 
& -73.21\%    \\
\hline\hline
\end{tabular}}
\label{tab: complexity}
\end{table}

\begin{table*}[t]
\centering
\small
\setlength{\abovecaptionskip}{1mm}
\setlength{\belowcaptionskip}{1mm}
\caption{
Quality comparisons are calculated on the Kodak~\cite{kodak2013} dataset. 
The experiment is conducted on an NVIDIA 3090 GPU. 
}
\begin{tabular}{l|cccccc} \hline\hline
\multirow{2}{*}{\textbf{Time Step}} & \multicolumn{6}{c}{\textbf{Kodak}} \\ \cline{2-7} 
& \textbf{PSNR} $\uparrow$ & \textbf{MS-SSIM} $\uparrow$ & \textbf{DISTS} $\downarrow$ & \textbf{LPIPS} $\downarrow$ & \textbf{FID} $\downarrow$ & \textbf{Inference Time (s)}   \\ \hline
Ours (step T = 5)   & 25.89 & 0.9037 & 0.0629 & 0.0961 & 34.79 & 1.9897\\
Ours (step T = 10)  & 25.91 & 0.9033 & 0.0642 & 0.0968 & 35.09 & 2.9424\\
Ours (step T = 20)  & 25.86 & 0.9026 & 0.0638 & 0.0970 & 34.42 & 5.1999\\
Ours (step T = 30)  & 25.84 & 0.9021 & 0.0623 & 0.0970 & 34.63 & 7.1554 \\
Ours (step T = 40)  & 25.82 & 0.9019 & 0.0618 & 0.0966 & 34.26 & 9.1428 \\
\rowcolor{cvprblue!10}
Ours (step T = 50) & 25.81 & 0.9013 & 0.0624 & 0.0972 & 34.09 & 11.1179 \\ 
Ours (step T = 100)  & 25.75 & 0.9003 & 0.0611 & 0.0973 & 33.39 & 21.1192\\
Ours (step T = 150)  & 25.71 & 0.8994 & 0.0611 & 0.0975 & 35.15 & 30.8623\\
Ours (step T = 200)  & 25.69 & 0.8988 & 0.0602 & 0.0977 & 33.74 & 41.0552\\\hline\hline
\end{tabular}
\label{tab: different time step}
\end{table*}

\noindent \textbf{Qualitative Comparisons.}
We provide qualitative comparisons to visually assess the performance of our compression method alongside eight state-of-the-art techniques. 
Fig.~\ref{fig: visual comparison} and ~\ref{fig: visual comparison 5} illustrate the results of our method in comparison to others constrained by the similar Bpp.
The output of our method showcases a comparison with other techniques, highlighting its ability to preserve fine details, especially in zoomed-in areas.
In contrast, the first three images demonstrate the output of the competing methods, which exhibits noticeable blurriness and loss of detail in the same region.

\noindent \textbf{Complexity Comparisons.}
Tab.~\ref{tab: complexity} summarizes the average encoding and decoding times for compared methods on the Kodak dataset. 
\textbf{Observation:} 
The table presents encoding and decoding times, along with BD-rate performance, for various methods evaluated on the Kodak dataset using an Nvidia GPU 3090. 
The encoding times vary across methods, with HiFiC taking 0.598 seconds.
Notably, our method demonstrates a competitive encoding time of 0.507 seconds.
\textbf{Analysis:} The results illustrate a clear trade-off between encoding and decoding times and compression performance. 
Methods like CRDR exhibit longer encoding and decoding times while only achieving a BD-rate performance of -13.19\%. 
In contrast, our method strikes a balance, providing both rapid encoding and decoding times alongside an impressive BD-rate performance (-73.21\%).

\subsection{Ablation Study}

\noindent \textbf{FFAB Block Performance Under Different Settings.}
In our experiments, the FFAB block aims to capture the low-frequency, high-frequency, vertical, and horizontal bands, which are highly correlated with the window size. 
\textbf{Observation:}
Fig.~\ref{fig: FFAB window size} analyzes the impact of different window sizes on compression performance, which is unique in the FFAB block design.
Herein, the Window 8 settings denote the following window sizes \(8*2 \times 8*2\), \(8/2 \times 8/2\), \(8*2 \times 8/2\), and \(8/2 \times 8*2\) to capture the isotropic and anisotropic features.
The Window 4 setting and Window 2 setting are also similar.
Our findings indicate that increasing the window size results in a decline in performance, as measured by PSNR and FID.
\textbf{Analysis:} 
Smaller window sizes allow the generative image compression model to focus on local features more effectively. 
This leads to a better capture of fine-grained details in decoded images, ultimately enhancing the overall compression performance of the proposed model.

\begin{figure*}[th]
\centering    
{\includegraphics[width=0.24\linewidth]{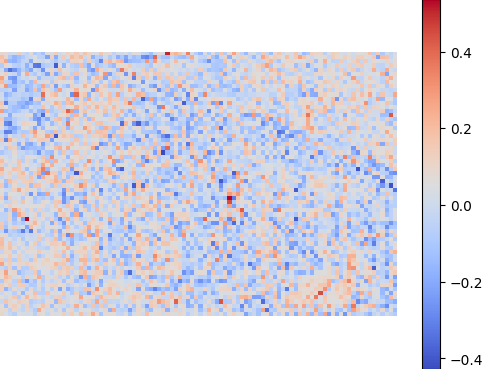}} \hskip1mm  
{\includegraphics[width=0.24\linewidth]{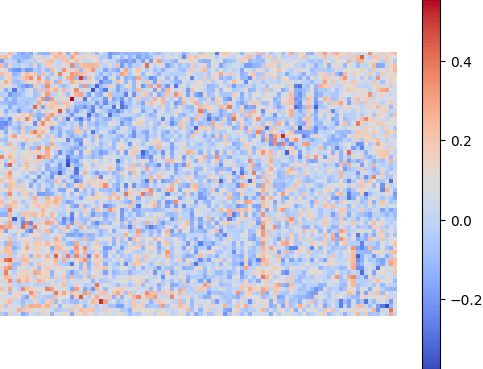}}\hskip1mm
{\includegraphics[width=0.24\linewidth]{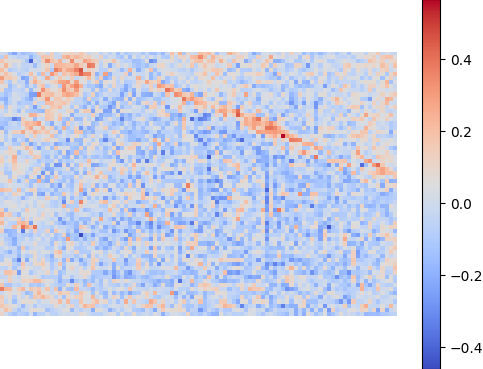}}\hskip1mm
{\includegraphics[width=0.24\linewidth]{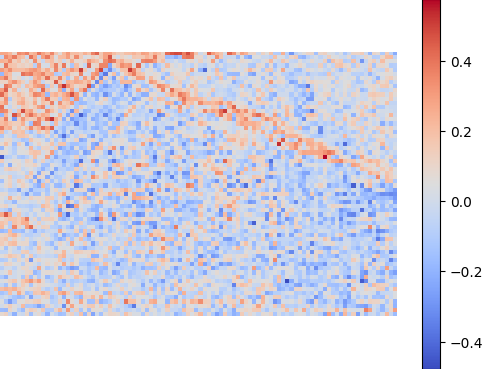}}\hskip1mm    
\vspace{3mm}

{\includegraphics[width=0.24\linewidth]{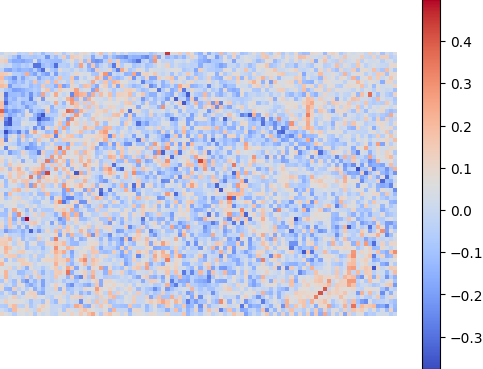}} \hskip1mm 
{\includegraphics[width=0.24\linewidth]{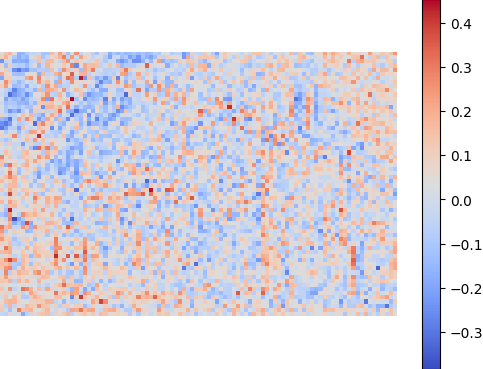}}\hskip1mm
{\includegraphics[width=0.24\linewidth]{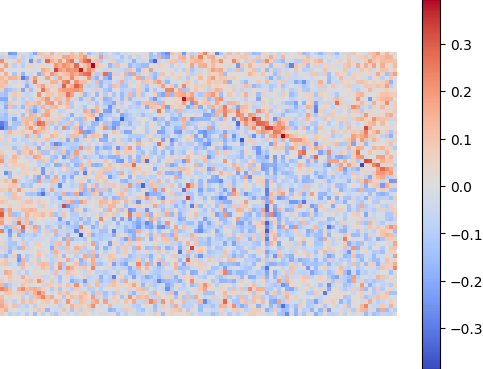}}\hskip1mm
{\includegraphics[width=0.24\linewidth]{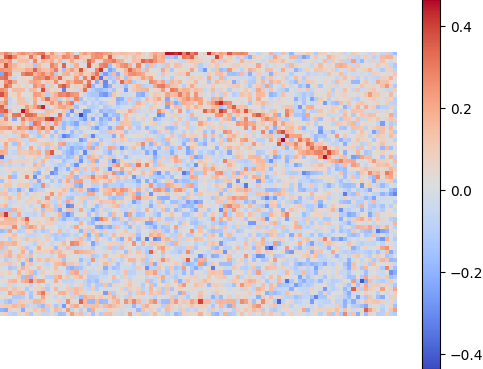}}\hskip1mm   
\vspace{3mm}

{\includegraphics[width=0.24\linewidth]{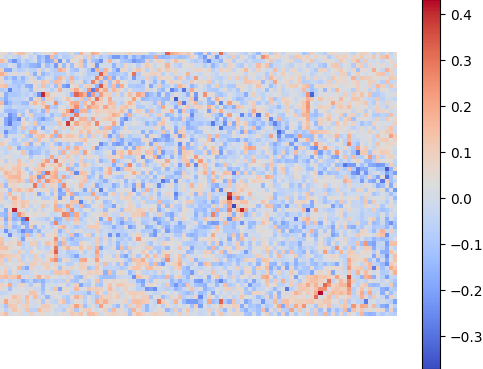}} \hskip1mm  
{\includegraphics[width=0.24\linewidth]{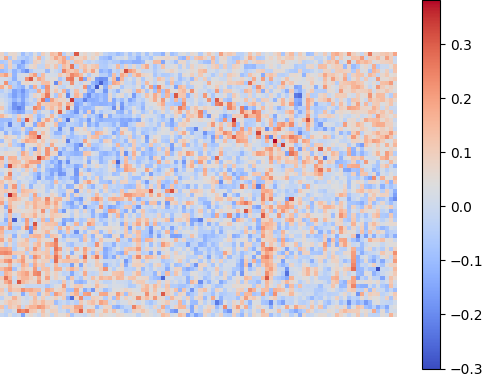}}\hskip1mm
{\includegraphics[width=0.24\linewidth]{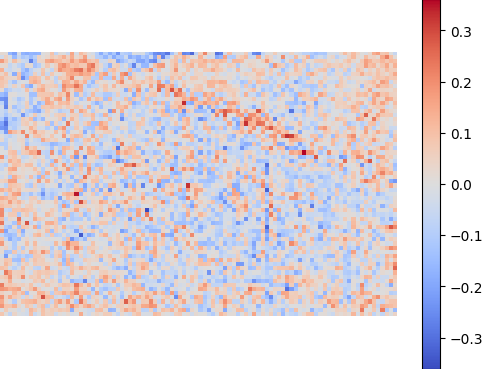}}\hskip1mm
{\includegraphics[width=0.24\linewidth]{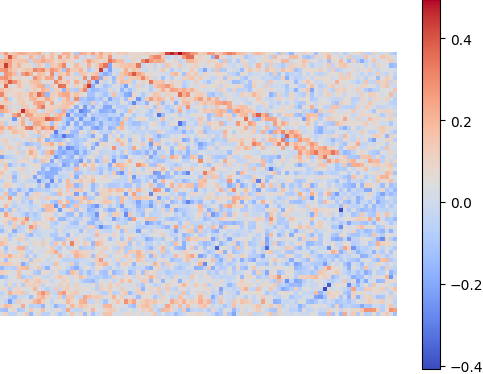}}\hskip1mm   
\caption{
The bpp for the first row is 0.1306, for the second row is 0.0992, and for the third row is 0.0595.
Low-frequency and high-frequency bands, as well as vertical and horizontal bands, are captured using window sizes of \(16 \times 16\), \(4 \times 4\), \(16 \times 4\), and \(4 \times 16\), respectively.
This visualization highlights varying bit rates across different areas, allowing us to observe the regions of emphasis.
}
\label{fig: FFAB performance under different rate 2}
\end{figure*}

\noindent \textbf{FFAB Performance under different rate.}
To further assess the effectiveness of the FFAB block performance under different rates, we present the results in Fig.~\ref{fig: FFAB performance under different rate 2} to visualize and support the contributions.
\textbf{Obseravation:}
The high bpp corresponds to the first set, suggesting that these images preserve more detail and information. 
The middle bpp indicates a reduction in detail, which may suggest a loss in clarity or information, potentially impacting the ability to distinguish features.
The low bpp reflects the least amount of detail, likely leading to a significant reduction in image quality and the visibility of features.
\textbf{Analysis:}
The adaptability of the FFAB block allows for improved texture fidelity across varying bpp levels. 
For example, at higher bpp values, the block can focus on recovering high-frequency details. 
In contrast, at lower bpp values, it can maintain a coherent representation of low-frequency components.
The designed FFAB block enables us to discern which textures require more attention based on their significance and texture complexity.

\noindent \textbf{Time Steps of Diffusion Model.}
We conduct the ablation study to determine the optimal number of time steps in our experiments, balancing inference time and decoded image quality, as shown in Tab.~\ref{tab: different time step}.
\textbf{Observation}: 
We could observe that as the number of time steps increases, both PSNR and MS-SSIM decrease, while DISTS also decrease, and FID and LPIPS metrics fluctuate.
\textbf{Analysis}: 
An increase in the number of time steps leads to gradual noise accumulation during the generation process. 
This accumulation can result in decreased structural similarity, negatively affecting PSNR and MS-SSIM, which are indicators of image quality. 
Conversely, the model may produce more visually appealing images at higher time steps, which could contribute to a lower DISTS value. 
Both FID and LPIPS reflect a trade-off between the diversity and quality of the generated images. 
As the number of time steps varies, the generated images may show fluctuations in both diversity and quality, resulting in unstable performance for these metrics.

\noindent \textbf{Effect of the Frequency Constraint.}
To evaluate the effectiveness of the frequency constraint in Eq.~\eqref{frequency_equ}, we remove this constraint and fine-tune the resulting model. 
On the Kodak dataset, the BD-rate gain in terms of DISTS decreases from -75.50\% to -72.41\%, while the FID gain decreases from -73.21\% to -71.11\%, using HiFiC as the anchor. 
These results demonstrate the effectiveness of the frequency constraint in improving both reconstruction fidelity and perceptual realism.

\section{Conclusion}
In this paper, we present a novel Fractal Frequency-Aware Band Image Compression (FFAB-IC) learning network, trained with frequency-aware loss supervision, aimed at enhancing generative image compression by leveraging a powerful generative prior. 
Our approach is motivated by two key aspects.
First, the reconstruction from content band representation closely aligns with frequency domain analysis, particularly in the context of image realism generation. 
To capitalize on this, we develop a fractal band learning network that performs frequency band feature operations, outperforming existing generative image compression methods across several common benchmarks.
Second, we incorporate learned contextual features from natural images in various directional contexts, extracted through our fractal band after window attention operations. 
This strategy effectively perceives rich and spatial information through texture feature representations, significantly improving visual quality.
Extensive experimental results validate the superiority of our method over previous approaches, highlighting the effectiveness of each component in the generative image compression process.

{\small
\bibliographystyle{IEEEtran}
\bibliography{reference}
}

\end{document}